\documentclass{l4dc2027}
\usepackage{microtype}

\usepackage[dvipsnames]{xcolor}
\usepackage{times}
\usepackage{graphicx}
\usepackage{booktabs}
\usepackage{float}
\usepackage[font=footnotesize]{caption}
\usepackage{mathtools,nicefrac,optidef}
\usepackage{algorithm,algpseudocode}

\newtheorem{assumption}{Assumption}
\newtheorem{problem}{Problem}

\newcommand{\cN}{\mathcal N}
\newcommand{\cF}{\mathcal F}
\newcommand{\E}{\mathbb E}
\newcommand{\KL}{D_{\mathrm{KL}}}
\newcommand{\Reg}{\operatorname{Reg}}

\title{Compute-Aware Submodular Online Learning \\ in Open Multi-Agent Systems\protect\thanks{\unexpanded{
			This work was supported in part by the China Scholarship Council (CSC).
}}}

\author{
	\Name{Jing Liu} \Email{y20220091@mail.ecust.edu.cn}\\
	\addr School of Mathematics\\
	East China University of Science and Technology, Shanghai, China
	\AND
	\Name{Luca Ballotta} \Email{luca.ballotta@unipd.it} \\
	\addr Department of Information Engineering\\
	University of Padova, Padova, Italy
	\AND
	\Name{Fangfei Li} \Email{lifangfei@ecust.edu.cn} \\
	\addr School of Mathematics and the Key Laboratory of Smart Manufacturing in Energy Chemical Process\\
	Ministry of Education, East China University of Science and Technology, Shanghai, China
	\AND
	\Name{Yang Tang} \Email{yangtang@ecust.edu.cn}  \\
	\addr Key Laboratory of Smart Manufacturing in Energy Chemical Process\\
	Ministry of Education, East China University of Science and Technology, Shanghai, China
	\AND
	\Name{Ruggero Carli} \Email{ruggero.carli@unipd.it}\\
	\addr Department of Information Engineering\\ 
	University of Padova, Padova, Italy}

\begin{document}
	
	\maketitle

\begin{abstract}
	This paper studies policy learning for cooperative task allocation in open multi-agent systems, where submodular utilities model diminishing marginal returns due to overlapping agent contributions.
	While standard methods evaluate every candidate action to update the policy,
	this may incur an excessive computational cost online.
	Hence, we impose a budget on these evaluations and analytically study the effect of query budget on online learning.
	Our algorithm samples a prescribed number of agent-action pairs and applies inverse-probability weighing to their marginal gains, yielding an unbiased gradient estimate of the expected stage utility for a Kullback-Leibler (KL) mirror-ascent update. 
	We derive a $1/2$-approximate regret bound that elegantly splits into a tracking error between optimal and learned policies, quantified by their KL divergence, and a query-dependent feedback term.
	Experiments on cooperative information coverage show that our approach retains high performance with significantly less computation online.
\end{abstract}


\section{Introduction}
\label{sec:intro}

Multi-agent task allocation refers to the problem where cooperative autonomous agents are dispatched to engage with given tasks.
Sub-additive agent contributions give rise to so-called  diminishing returns: each agent's individual (or marginal) contribution to global performance decreases when more agents engage with a target.
This is common in robotics where diminishing returns model, e.g., overlapping camera fields-of-view in cooperative coverage~\citep{Tzoumas16acc-sensorPlacementKalman}.
Submodularity offers an elegant framework to formally
capture such overlaps and derive provable suboptimality gaps if the feasible set has added structure in the form of a matroid, which includes common cardinality and partition  constraints~\citep{nemhauser1978analysis,calinescu2011maximizing}.

The intersection between submodularity and data-driven coordination policies is still underexplored, especially in multi-agent systems.
Online coordination methods use knowledge about submodular marginal gains~\citep{xu2023online,zhang2025nearoptimal,zhang2025effective}, while bandit
methods leverage feedback on selected actions~\citep{zhang2019online,wan2023bandit,xu2023bandit}.
Related to submodular data-driven optimization,
previous works by the authors study submodular multi-agent reinforcement learning;
\cite{liu2026submapg} designs policy gradient tailored to submodular rewards,
while~\cite{liu2026submapl} learns tabular softmax policies via Kullback-Leibler (KL) mirror ascent.

While these studies leverage submodularity to evaluate the  marginal gains of alternative actions and iteratively improve the policy,
they assume no constraint on the learning agent feedback,
i.e., all candidate actions can be evaluated at each policy update.
However, this assumption may fail in online learning; if the physical learning device (e.g., a mobile robot or smart sensor) carries limited compute,
it cannot timely evaluate all marginal gains corresponding to a large number of actions or agents,
introducing a latency bottleneck for real-time operation.
Moreover, open systems, where agents may join or leave during
operation~\citep{deplano2026optimization},
exacerbate this issue;
each agent must select its action while accounting for possible changes in the team.

Budgeted feedback has been studied in label-efficient prediction~\citep{cesabianchi2005label}, reinforcement learning~\citep{efroni2021confidence}, and nonstationary bandits~\citep{li2025constrained}. 
However, these works limit access to labels or rewards rather than counterfactual marginal evaluations used to update the policies,
while we focus on the latter issue which is central to compute-aware learning.

\paragraph{Contribution.}
To address this challenge, we extend SubMAPL~\citep{liu2026submapl} to online learning with budgeted marginal feedback.
Instead of evaluating every candidate action, we impose a given number of marginal queries to estimate the gradient of the expected utility.
This reduces the number of marginal evaluations at each round while allowing the policy to adapt as the agents and feasible actions change.
Our contributions are summarized as follows.
\begin{itemize}
	\item We propose a policy-aware query allocation strategy that distributes a given number of marginal evaluations across agent-action pairs at each learnig round. Unlike methods that restrict reward observations~\citep{efroni2021confidence,li2025constrained}, we allocate counterfactual marginal queries for policy updates. In particular, we derive and characterize an optimal allocation rule that minimizes an upper bound on the second moment of the  policy gradient estimator.
	\item We develop an online submodular policy learning method with a given number of queries at each round.
	Our method constructs an unbiased gradient estimator using only the selected marginal evaluations, enabling online learning with limited compute in real time.
	A fixed-size sampling procedure ensures that the query budget is always satisfied, updating policies online as agents and feasible actions change.
	\item We establish a $1/2$-approximate dynamic regret bound for our method. Compared with recent studies on multi-agent online learning~\citep{zhang2025nearoptimal,zhang2025effective}, our bound highlights the distinct contributions of feedback budget and of time-varying agent set due to open systems.
	Numerical experiments on cooperative coverage showcase performance close to the compute-unaware full-feedback baseline despite a substantial reduction in marginal queries.
\end{itemize}

\paragraph{Organization.}
We introduce the multi-agent submodular online learning problem in \autoref{sec:problem}.
In \autoref{sec:prelim},
we recap SubMAPL,
which we extend to budgeted  online learning in \autoref{sec:method}.
In \autoref{sec:analysis}, we establish a provable regret bound and propose a strategy to sample counterfactual feedback that minimizes it.
We report simulations results in \autoref{sec:experiments} and conclude in \autoref{sec:conclusion}.

\section{Setup and Problem Statement}
\label{sec:problem}

\subsection{Multi-Agent Task Allocation Setup}
\label{sec:setup}

The system evolves across rounds $t=1,\ldots,T$.
State $s_t$ includes agent configurations, tasks, and any other variables required to evaluate performance.
Let $\mathcal N_t$ denote the set of active agents at round $t$. 
Each agent $i\in\mathcal N_t$ selects one action from a finite, potentially time-varying set $\mathcal A_{i,t}$.
We define the total number of actions at round $t$ as $d_t=\sum_{i\in\mathcal N_t}m_{i,t}$ with $m_{i,t}=|\mathcal{A}_{i,t}|$.
The agent-action ground set $
\Omega_t=\{(i,a):i\in\mathcal N_t,\ a\in\mathcal A_{i,t}\}$ gathers all possible agent-action pairs at round $t$.
A joint action $A_t\subseteq\Omega_t$ contains one action per active agent and $A_t^{-i}=A_t\setminus\{(i,a_{i,t})\}$ excludes agent $i$.

Let $F_t(A;s_t)$ denote the team utility when the joint action $A$ is applied at state $s_t$.
For an element $e\in\Omega_t$, we denote its marginal gain (relative to $A$) as $F_t(e\mid A;s_t)=F_t(A\cup\{e\};s_t)-F_t(A;s_t)$. A set function is submodular if $F_t(e\mid B;s_t)\ge F_t(e\mid C;s_t)$ for all $B\subseteq C\subseteq\Omega_t$ and $e\in\Omega_t\setminus C$. Submodularity captures diminishing marginal returns associated with overlapping, i.e., sub-additive agent contributions to team performance~\citep{nemhauser1978analysis}.

\begin{assumption}[Stage utility]
	\label{ass:stage}
	Each $F_t(\cdot;s_t):2^{\Omega_t}\to\mathbb{R}$ is normalized, monotone, and submodular.
	There exists a constant $L>0$ such that $0\le F_t((i,a)\mid A^{-i};s_t)\le L$
	for every $(i,a)\in\Omega_t$ and every collection $A^{-i}$ containing one action from each active agent other than $i$.
\end{assumption}

During online operation,
a learner samples joint actions from a categorical policy over active agents.
Let $x_{i,a}$ denote the probability that agent $i$ executes action $a$.
The set of policies at round $t$ is
\begin{equation}\label{eq:face}
	\cF_t=\left\{x\in\mathbb R_+^{|\Omega_t|}:\sum_{a\in\mathcal A_{i,t}}x_{i,a}=1,\ \forall i\in\mathcal N_t\right\}
\end{equation}
and the policy of each agent $i$ is the categorical probability distribution $\{x_{i,a}\}_{a\in\mathcal{A}_{i,t}}$. 
For the multi-agent policy $x_t\in\cF_t$, let $\mathcal D_t(x_t)$ denote the distribution over joint actions induced by the categorical distributions of active agents.
The conditional expected stage utility is
\begin{equation}
	\widetilde f_t(x_t;s_t)=\E_{A\sim\mathcal D_t(x_t)}[F_t(A;s_t)\mid s_t].
	\label{eq:pme}
\end{equation}

Our objective is to learn policies $x_t$ online that achieve high stage utility despite changes in the state, active agents, and feasible actions that induce time-varying utilities.
As a performance metric, following common formulations in online learning,
we compare the learned policy with a stagewise optimal policy evaluated at the state visited by the learner at each round~\citep{xu2023online}.

\begin{problem}[Regret-optimal online learning]\label{prob:main-regret}
	Let an optimal time-varying policy be 
	\begin{equation}\label{eq:xstar}
		x_t^\textnormal{OPT}\in\arg\max_{x_t\in\cF_t} \widetilde f_t(x_t;s_t) \quad t=1,\dots,T.
	\end{equation}
	Define the $1/2$-approximate dynamic regret of a time-varying policy $x=\{x_t\}_{t=1}^T\in\cF=\cF_1\times\dots\times\cF_T$ as 
	\begin{equation}
		\label{eq:dynamic-regret}
		\Reg_T^{1/2}(x)=\E_{\tau}\left[\sum_{t=1}^T\left(\frac12\widetilde f_t(x_t^\textnormal{OPT};s_t)-\widetilde f_t(x_t;s_t)\right) \Big| s_1\right]
	\end{equation}
	where $\tau$ denotes the stochastic state trajectory induced by the multi-agent policy.
	We aim to learn a regret-optimal policy $x^*=\{x_t^*\}_{t=1}^T$ that satisfies
	\begin{equation}
		\label{eq:prob-mininize-dynamic-regret}
		x^*\in\arg\min_{x\in\cF}\Reg_T^{1/2}(x).
	\end{equation}
\end{problem}

We use $1/2$-approximation instead of classic regret due to submodularity; analytical intuition is given in \autoref{sec:prelim}.
In the next section, we recap a policy gradient method tailored to submodular utilities and its computational limitation, which we adapt in \autoref{sec:method} to solve Problem~\ref{prob:main-regret}.

\section{Preliminaries: Submodular Policy Gradient}
\label{sec:prelim}

In paper~\citep{liu2026submapl},
the authors propose a KL mirror-ascent policy gradient method for multi-agent reinforcement learning with tabular softmax policies, SubMAPL,
which can be readily adapted to tackle Problem~\ref{prob:main-regret}.
Central to the formal guarantees of this method,
the partition multilinear extension (PME) $\widetilde{f}$ of a submodular set function $F$ is a continuous relaxation of $F$ defined on the extended policy domain $\mathcal P_t=\{x\in\mathbb R_+^{|\Omega_t|}:\sum_{a\in\mathcal A_{i,t}}x_{i,a}\le1,\ \forall i\in\mathcal N_t\}$,
where $x_{i,a}$ has the meaning described in \autoref{sec:setup}: 
for $x\in\mathcal P_t$, agent $i$ selects action $a$ with probability $x_{i,a}$ and selects no action with the remaining probability.
Notice that an arbitrary element in $\mathcal{P}_t$ is \emph{not} a proper multi-agent policy since some agent's categorical distributions may not be normalized to $1$.
By defining the restriction to the so-called categorical face $\cF_t=\partial\mathcal{P}_t$ defined in~\eqref{eq:face},
it is possible to prove that the PME $\widetilde{f}_t(x_t;s_t)$ of the stage utility $F_t(A;s_t)$ coincides with its expectation as per~\eqref{eq:pme} and that its gradient satisfies
\begin{equation}
	\nabla_{i,a}\widetilde f_t(x;s_t)=\E_{A^{-i}\sim\mathcal D_t^{-i}(x)}\left[F_t((i,a)\mid A^{-i};s_t)\right],
	\label{eq:pme-gradient}
\end{equation}
where $\mathcal D_t^{-i}(x)$ denotes the distribution of active agents' actions other than $i$~\citep{liu2026submapl}.

Given a joint action $A_t\sim\mathcal D_t(x_t)$, the counterfactual-gain feedback associated with $(i,a)\in\Omega_t$ evaluates the marginal gain of the agent-action pair $(i,a)$ as
\begin{equation}
	h_{i,a,t}=F_t((i,a)\mid A_t^{-i};s_t).
	\label{eq:local-marginal}
\end{equation}
Then,
for a state $s_t$ and policy $x_t$, this feedback provides an unbiased estimator of the corresponding PME gradient coordinate, since combining~\eqref{eq:pme-gradient} and~\eqref{eq:local-marginal} readily yields
\begin{equation}
	\E_{A^{-i}\sim\mathcal D_t^{-i}(x)}\left[h_{i,a,t}\right]=\nabla_{i,a}\widetilde f_t(x_t;s_t).
	\label{eq:marginal-gradient-link}
\end{equation}
Equivalently,
the counterfactual-gain vector $h_t=(h_{i,a,t})_{i,a}$ is a \emph{stochastic gradient} of $\widetilde{f}_t(x_t;s_t)$.
SubMAPL updates the agents' categorical policies through KL mirror ascent,
where the policy probabilities are assumed to satisfy the following condition.
\begin{assumption}[Policy support]
	\label{ass:support}
	For every round $t$, it holds $x_{i,a,t}>0$ for all $a\in\mathcal A_{i,t},i\in\mathcal N_t$.
\end{assumption}
Given the full counterfactual-gain vector $h_t$ and stepsize $\eta>0$, the KL mirror-ascent update is
\begin{equation}
	\bar x_{t+1}=\mathrm{KL}(x_t)=\arg\max_{x\in\cF_t}\left\{\eta\langle h_t,x\rangle-\KL(x\Vert x_t)\right\}
	\label{eq:kl-mirror-step}
\end{equation}
where $\KL$ denotes the standard Kullback-Leibler divergence~\citep{beck2003mirror}.
This update admits the following closed-form expression for each active agent $i$:
\begin{equation}
	\bar x_{i,a,t+1}=\frac{x_{i,a,t}\exp(\eta h_{i,a,t})}{\sum_{b\in\mathcal A_{i,t}}x_{i,b,t}\exp(\eta h_{i,b,t})}.
	\label{eq:mirror-update}
\end{equation}
We denote the updated policy on the current domain by $\bar x_{t+1}\in\cF_t$,
which does not consider changes in time-varying agent and action sets.
The policy actually implemented in the next round is $x_{t+1}=\mathcal M_{t+1}(\bar x_{t+1})\in\cF_{t+1}$,
where map $\mathcal M_{t+1}$ adjusts tabular policy parameters to accommodate potentially different agent set $\mathcal{N}_{t+1}$ and actions sets $\mathcal{A}_{i,t+1}$ at the next round.
We omit its formal definition for the sake of readability and refer the reader to~\cite{liu2026submapl} for details.

SubMAPL learns tabular softmax policies.
The learner maintains agent-indexed logits $\theta_i(o,a)$ for every observation-action pair.
Given observation $o_{i,t}$ of $s_t$, the policy of agent $i$ is
\begin{equation}
	x_{i,a,t}=\frac{\exp(\theta_i(o_{i,t},a))}{\sum_{b\in\mathcal A_{i,t}}\exp(\theta_i(o_{i,t},b))}.
\end{equation}
Then,
the update in~\eqref{eq:mirror-update} is implemented by adding $\eta h_{i,a,t}$ to current logits.
Finite logits and feasibility masking through migration $\mathcal{M}$ preserve the positive support required by Assumption~\ref{ass:support}.

SubMAPL can be readily used to learn a time-varying multi-agent policy $x_t$ online,
in agreement with Problem~\ref{prob:main-regret}.
However,
a computational bottleneck arises:
update~\eqref{eq:mirror-update} requires querying the counterfactual marginal gain $h_{i,a,t}$ for each active agent-action pair $(i,a)$ at each round $t$.
The time complexity of this calculation does not (jointly) scale with the number of agents and actions,
clashing against real-time requirements for large systems.
Hence,
building on the framework of SubMAPL,
in \autoref{sec:method} we introduce a computationally controlled budgeted gradient estimator that replaces $h_t$ while preserving the KL mirror-ascent update and policy-migration mechanism.

\section{Policy Learning With Budgeted Queries}
\label{sec:method}

The KL update in \eqref{eq:mirror-update} requires evaluating a \emph{marginal query} for every active agent-action pair $(i,a)\in\Omega_t$,
i.e., evaluating the counterfactual gain $h_{i,a,t}$ in \eqref{eq:local-marginal} holding the other agents' actions $A_t^{-i}$ fixed.
Next, we construct an unbiased estimator of $\nabla\widetilde{f}_t$ with a fixed budget on counterfactual queries.

At each round $t$, the learner selects $Q_t\in\{1,\ldots,d_t\}$ queries from the active agent-action pairs,
where we recall that $d_t$ is the total number of actions for all agents at time $t$.
Let $Z_{i,a,t}\in\{0,1\}$ be equal to $1$ if $(i,a)\in\Omega_t$ is queried and zero otherwise.
Then, given original SubMAPL iterates and budget constraint,
we formally define the problem tackled in the remaining.
\begin{problem}[Budgeted submodular online learning]
	\label{prob:budgeted-online}
	Learn a regret-optimal policy $x_t$ via SubMAPL subject to a budget $Q_t$ on marginal feedback queries:
	\begin{mini!}
		{x\in\cF}{\Reg_T^{1/2}(x)}
		{\label{eq:stagewise-optimization}
		}
		{}
		\addConstraint{x_{t+1}}{=\mathcal{M}_{t+1}(\mathrm{KL}(x_{t}))}{\quad\forall t=1,\dots,T}
		\addConstraint{\sum_{(i,a)\in\Omega_t}Z_{i,a,t}}{ =Q_t}{\quad\forall t=1,\dots,T\label{eq:query-constraint}}
		\addConstraint{Z_{i,a,t}}{ \in\{0,1\}.}{\quad\forall (i,a)\in\Omega_t,\forall t=1,\dots,T}
	\end{mini!}
\end{problem}

In the following,
we tackle Problem~\ref{prob:budgeted-online} by designing a stochastic querying mechanism that is guaranteed to satisfy the budget constraint and returns an unbiased gradient estimator for $\widetilde{f}_t$ at each round $t$.
Denote by $\mathcal I_t^-=\{s_t,\mathcal{N}_t,x_t,Q_t\}$ the information available before the current marginal gains are observed:
this is all and only information available to select queries.
We define the inclusion probability $\rho_{i,a,t}=\Pr(Z_{i,a,t}=1\mid\mathcal I_t^-)$ as the chance that pair $(i,a)$ is queried at round $t$.
Conditioned on $\mathcal I_t^-$,
query randomization is independent of current utility and sampled joint action.
To ensure that the construction is well posed,
we impose a positive lower bound $\rho_{\min,t}\ge\rho_{\min}>0$ on the inclusion probabilities.
Thus,
query design satisfies
\begin{align}
	&\rho_{\min,t}\le\rho_{i,a,t}\le1 \quad \forall\;(i,a)\in\Omega_t
	\label{eq:rho-range}\\
	&\sum_{(i,a)\in\Omega_t}\rho_{i,a,t}=Q_t
	\label{eq:rho-budget}\\
	&\sum_{(i,a)\in\Omega_t}Z_{i,a,t}=Q_t.
	\label{eq:hard-budget}
\end{align}
Feasibility is guaranteed when $Q_t\ge d_t\rho_{\min,t}$.
Using the queried marginal gains,
we construct the inverse-probability PME gradient coordinate estimator as
\begin{equation}
	\widehat g_{i,a,t}=\frac{Z_{i,a,t}}{\rho_{i,a,t}}h_{i,a,t}.
	\label{eq:sparse-estimator}
\end{equation}
The estimated gradient $\widehat g_t=(\widehat g_{i,a,t})_{(i,a)}$ is then used in the KL mirror-ascent update~\eqref{eq:kl-mirror-step} in place of $h_t$. Note that $\widehat g_{i,a,t}=0$ if $(i,a)$ is not queried: its marginal feedback is not used in the current update, which does not imply that the corresponding PME gradient coordinate is zero. 
As we prove later, this estimator is indeed unbiased for the PME gradient $\nabla\widetilde{f}_t$ after averaging over joint-action and query sampling, formally supporting our online learning algorithm and its regret analysis.

\paragraph*{Deterministic query sampling via rounding.}
To deterministically enforce the query budget from probabilistic querying,
we apply randomized pipage rounding,
a classic pairwise-rounding procedure to satisfy cardinality constraints.
Given inclusion vector $\rho_t=(\rho_{i,a,t})_{(i,a)}$,
the procedure generates a selection vector $Z_t\in\{0,1\}^{d_t}$ satisfying
$\sum_{(i,a)\in\Omega_t}Z_{i,a,t}=Q_t$ and $\E[Z_{i,a,t}\mid\mathcal G_t]=\rho_{i,a,t} \ \forall (i,a)$.
The conditional inclusion probabilities are preserved by performing query rounding independently of joint-action sampling.
For details, we refer the reader to~\citep{gandhi2006dependent}.

Tabular softmax policy parametrization and migration for open system dynamics follow the mechanism of SubMAPL described in \autoref{sec:prelim}.
The final algorithm with budgeted gradient estimation is summarized in \autoref{alg:query-submapl}.
Note that pipage rounding requires at most $d_t-1$ pairwise updates.
When $Q_t=d_t$, all pairs are queried and the update coincides with the full marginal-feedback KL update of SubMAPL.
In the next section,
we analytically bound the approximate dynamic regret of \autoref{alg:query-submapl} and propose a strategy to choose inclusion probabilities $\rho_t$ based on that.

\begin{algorithm}[t]
	\caption{Online policy learning with query-budgeted SubMAPL updates}
	\label{alg:query-submapl}
	\begin{algorithmic}[1]
			\Require Current policy $x_t\in\cF_t$, budget $Q_t$, lower bound $\rho_{\min,t}$, step size $\eta$
			\State Observe $s_t$, $\mathcal N_t$, and $\{\mathcal A_{i,t}\}_{i\in\mathcal N_t}$
			\State Select admissible inclusion probabilities $\rho_t$
			\State Sample $Z_t$ by fixed-size dependent rounding of $\rho_t$
			\State Sample $A_t\sim\mathcal D_t(x_t)$ independently of query rounding
			\State Evaluate $h_{i,a,t}$ for the pairs with $Z_{i,a,t}=1$ at the fixed state $s_t$
			\State Execute $A_t$ and observe the next state
			\State Construct $\widehat g_t$ using \eqref{eq:sparse-estimator}
			\State Update $\bar x_{t+1}$ using \eqref{eq:mirror-update} with $h_t$ replaced by $\widehat g_t$
			\State Set $x_{t+1}=\mathcal M_t(\bar x_{t+1})$ on the next active domain
		\end{algorithmic}
\end{algorithm}

\section{Performance Analysis}
\label{sec:analysis}

In this section, we establish a provable bound on approximate regret for \autoref{alg:query-submapl} that holds uniformly across allocations $\rho_t$.
Then,
we propose an optimization-based allocation $\rho^\star$ that minimizes our regret bound; hence, while it need not be optimal w.r.t. the original Problem~\ref{prob:main-regret}, it provides a principled way to query counterfactual marginals under budget constraint.
Proofs are in Appendix.

To account for changes in the stagewise comparator and policy domain, define the policy-dependent open-system KL tracking variation~\citep{liu2026submapl} as
\begin{equation}
	\mathcal V_T^{\mathrm{open}}=\E\!\Bigg[\sum_{t=1}^{T-1}\Big(\KL(x_{t+1}^\star\Vert x_{t+1}) - \KL(x_t^\star\Vert\bar x_{t+1})\Big)_+\Bigg],
	\label{eq:open-variation}
\end{equation}
where $(z)_+=\max\{z,0\}$. Each KL divergence compares policies on the same active agent--action domain; $\bar x_{t+1}$ and $x_{t+1}$ are the pre- and post-migration policies defined in Section~\ref{sec:method}.

\begin{theorem}[Query-dependent regret bound]
	\label{thm:query-regret}
	Under Assumptions~\ref{ass:stage}--\ref{ass:support}, Algorithm~\ref{alg:query-submapl} satisfies
	\begin{equation}
		\Reg_T^{1/2}\le\frac{\E[\KL(x_1^\star\Vert x_1)]+\mathcal V_T^{\mathrm{open}}}{2\eta}
		+\frac{(e-2)\eta L^2}{2}\sum_{t=1}^T\E\!\left[\sum_{(i,a)\in\Omega_t}\frac{x_{i,a,t}}{\rho_{i,a,t}}\right].
		\label{eq:main-regret}
	\end{equation}
\end{theorem}

As a baseline, we specialize this general bound to a naive allocation strategy that samples counterfactual marginals uniformly at random across active agent-action pairs.

\begin{corollary}[Regret bound with uniform querying]
	\label{cor:budget-forms}
	Under Assumptions~\ref{ass:stage}--\ref{ass:support},
	let \autoref{alg:query-submapl} query $q_{i,t}$ actions uniformly for each agent, i.e., $q_{i,t}\sim U([1,m_{i,t}])$ for all $i\in\mathcal{N}_t$ and $t$.
	It holds
	\begin{equation}
		\Reg_T^{1/2}\le\frac{\E[\KL(x_1^\star\Vert x_1)]+\mathcal V_T^{\mathrm{open}}}{2\eta}
		+\frac{(e-2)\eta L^2}{2}\sum_{t=1}^T\E\!\left[\sum_{i\in\mathcal N_t}\frac{m_{i,t}}{q_{i,t}}\right]
		\label{eq:uniform-regret}
	\end{equation}
	where the KL tracking variation is the one induced by uniform querying.
\end{corollary}

\paragraph*{Policy-aware query allocation.}
The query-dependent regret bound in Theorem~\ref{thm:query-regret} motivates a strategic selection of inclusion probabilities to reduce regret of the learned policy.
From~\eqref{eq:main-regret}, the policy-weighted estimator second moment is bounded by $L^2\Phi_t(\rho_t)$ with
\begin{equation}
	\Phi_t(\rho_t)=\sum_{(i,a)\in\Omega_t}\dfrac{x_{i,a,t}}{\rho_{i,a,t}}.
\end{equation}
Consequently,
we determine the inclusion probabilities by solving
\begin{mini!}
	{\rho}
	{\Phi_t(\rho)}
	{\label{eq:allocation-program}}
	{}
	\addConstraint{\rho_{\min,t}\le\rho_{i,a,t}}{\le1}{\quad\forall\,(i,a)\in\Omega_t,\forall \,t=1,\dots,T}
	\addConstraint{\sum_{(i,a)\in\Omega_t}\rho_{i,a,t}}{=Q_t}{\quad\forall \,t=1,\dots,T.}
\end{mini!}
The resulting allocation depends only on the current policy and query budget and can therefore be computed before observing the marginal gains.

\begin{proposition}[Regret-optimal query allocation]
	\label{prop:allocation}
	Under Assumption~\ref{ass:support}, if $d_t\rho_{\min,t}\le Q_t$, the optimization problem \eqref{eq:allocation-program} admits a unique minimizer given by		$\rho_{i,a,t}^{\star}=\operatorname{clip}_{[\rho_{\min,t},1]}\!\left(\tau_t\sqrt{x_{i,a,t}}\right),$
	where $\operatorname{clip}_{[l,u]}(z)=\min\{u,\max\{l,z\}\}$ and $\tau_t>0$ is chosen so that $\sum_{(i,a)\in\Omega_t}\rho_{i,a,t}^{\star}=Q_t$.
	In particular, $\tau_t$ need not be unique if all coordinates are clipped.
	When no clipping is active, it holds
	\begin{equation}\label{eq:unclipped-rho}
		\rho_{i,a,t}^{\star}=\frac{Q_t\sqrt{x_{i,a,t}}}{\sum_{(j,b)\in\Omega_t}\sqrt{x_{j,b,t}}},
		\qquad
		\Phi_t(\rho_t^\star)=\frac{\left(\sum_{(i,a)\in\Omega_t}\sqrt{x_{i,a,t}}\right)^2}{Q_t}.
	\end{equation}
\end{proposition}

\begin{corollary}[Regret bound with policy-aware querying]\label{cor:regret}
	Let $\rho_t=\rho_t^\star \ \forall t=1,\dots,T$.
	It holds
	\begin{equation}
		\Reg_T^{1/2}\le\frac{\E[\KL(x_1^\star\Vert x_1)]+\mathcal V_T^{\mathrm{open}}}{2\eta}
		+\frac{(e-2)\eta L^2}{2}\sum_{t=1}^T\E\left[\frac{d_t|\mathcal N_t|}{Q_t}\right].
		\label{eq:optimal-regret}
	\end{equation}
\end{corollary}

To characterize the benefit of optimization~\eqref{eq:allocation-program},
we compare with naive uniform querying by defining the \emph{effective action dimension}
\begin{equation}
	d_{\mathrm{eff}}(x_t)=\frac{\left(\sum_{(i,a)\in\Omega_t}\sqrt{x_{i,a,t}}\right)^2}{|\mathcal N_t|}.
	\label{eq:effective-dimension}
\end{equation}
By \eqref{eq:effective-dimension}, blockwise normalization and the Cauchy--Schwarz inequality yield $|\mathcal N_t|\le d_{\mathrm{eff}}(x_t)\le d_t$.
Under uniform allocation $\rho^{\mathrm{unif}}_{i,a,t}=\nicefrac{Q_t}{d_t}$, the unclipped optimum $\rho_{i,a,t}^{\star}$ satisfies
\begin{equation}
	\frac{\Phi_t(\rho_t^\star)}{\Phi_t(\rho_t^{\mathrm{unif}})}=\frac{d_{\mathrm{eff}}(x_t)}{d_t},
	\label{eq:effective-dimension-ratio}
\end{equation}
which characterizes the second-moment improvement over uniform querying with inactive clipping.

Theorem~\ref{thm:query-regret} and Corollary~\ref{cor:regret} separate the KL tracking variation from the query-dependent feedback term.
Establishing sublinear regret requires suitable bounds on $\mathcal V_T^{\mathrm{open}}$ and $\sum_{t=1}^T\E[d_t\mathcal{N}_t]$.

\section{Experiments}
\label{sec:experiments}

\begin{figure}
	\centering
	\includegraphics[width=0.92\textwidth]{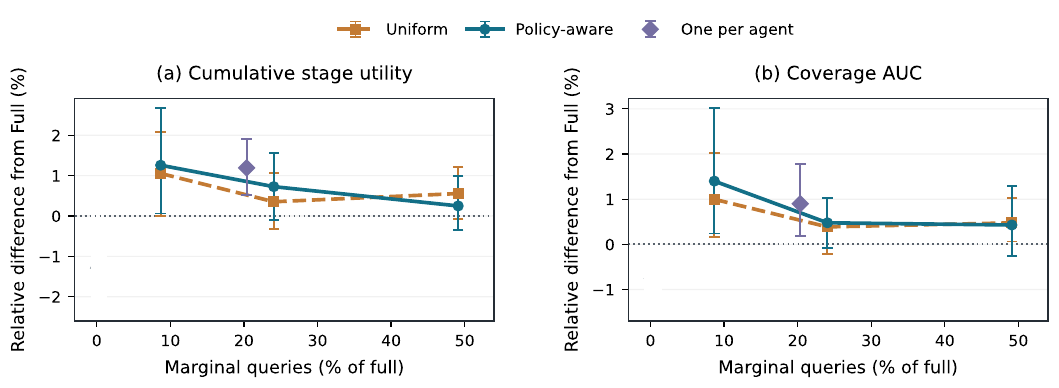}
	\caption{Relative cumulative stage utility (a) and coverage AUC (b). Marks and bars respectively show means and $95\%$ confidence intervals across cell-weight fields, agent schedules, and seeds.}
	\label{fig:frontier}
\end{figure}

We study how query budget supports online policy learning in open systems where agents enter and leave overtime.
All reported performance estimates use 20 held-out seeds.

\subsection{Cooperative coverage task}
We consider $|\mathcal{N}_t|\le\max_t|\mathcal{N}_t|=5$ agent on a $30\times30$ grid through $T=200$ rounds.
An action keeps the agent still or moves it by one cell in a cardinal direction,
such that all possible resulting positions cover a $3\times3$ footprint $C_{i,a,t}$.
Each cell $c$ has weight $w_c\ge0$ satisfying $\sum_c w_c=1$, quantifying, e.g., sensing information gathered at the cell.
We denote the cells covered by round $t$ as $U_t$ and set the team's utility at round $t$ as the newly covered weighted cells (satisfying Assumption~\ref{ass:stage}):
\begin{equation}
	F_t(A_t;s_t)=\sum_{c\in\bigcup_{(i,a)\in A_t}C_{i,a,t}\setminus U_t}w_c.
\end{equation}
The learner learns policies online using \autoref{alg:query-submapl}.
If an agent (temporarily) leaves the system, the relevant policy logits are kept frozen.
Observations contain workspace sector, local covered/value categories, and previous action,
while unseen observation are assigned uniform action probabilities.

We average results across three cell-weight fields, namely uniform, Gaussian mixture with two components, and log-Gaussian,
and three agent schedules, namely static (closed) system with $|\mathcal{N}_t|\equiv5$, a ``ramp schedule'' with $2,4,5,4,2$ agents changing at rounds $50,100,150,175$, and a ``shock'' schedule with $5,2,5$ agents changing at rounds $70,130$. We use the same environment seeds and learning rate $\eta=2$ for every condition and run each of them with $20$ random seeds.

Full feedback queries all $d_t\le\max_{i}|\mathcal{N}_t| m_{i,t}=45$ agent-action pairs at each round $t$, whereby each agent can execute at most $\max_{i}m_{i,t}=9$ actions, which corresponds to the original unconstrained implementation with full stochastic gradient $h_t$.
We compare optimized and uniform querying with budget $Q_t=\operatorname{round}(\beta d_t)$ for $\beta\in\{0.10,0.25,0.50\}$ and lower bound $\rho_{\min,t}=0.1\nicefrac{Q_t}{d_t}$.
In addition,
we compare a uniform allocation that draws one candidate action for each active agent at each round,
whose budget therefore need not match the other budgeted querying policies.
Queries are selected before evaluating counterfactual marginals,
hence making querying causal and practically meaningful under online learning in real time applications.
We compare policies based only on the number of queries; deeper comparison on actual compute cost, e.g., CPU cycles or wall-clock time, will be investigated in future work.

\subsection{Average Performance Versus Query Settings}

\autoref{fig:frontier} reports relative cumulative stage utility and the area-under-curve coverage metric  $\mathrm{AUC}=T^{-1}\sum_{t=1}^{T}\sum_{s=1}^{t}F_s(A_s;s_s)$,
w.r.t. full-feedback feedback, i.e., no budget constraint,
averaged across all cell-weight fields, agent schedules, and seeds.
Formally, a method's relative difference is $100(\nicefrac{V}{V_{\mathrm{full}}}-1)$, where $V$ is either metric and $V_{\mathrm{full}}$ is its corresponding full-feedback value. 
AUC captures coverage throughout an episode, including differences before final coverage saturates.
Bars mark percentile $95\%$ confidence intervals.

All budgeted methods perform close to the full-query baseline,
with percentage displacement within $2\%$ in most cases.
For instance,
using about $10\%$ of marginal queries (corresponding to $\beta=0.10$), the two budgeted methods feature mean AUC differences of $1.4\%$ for policy-aware allocation and slightly below $1\%$ for uniform allocation.
In fact,
it is remarkable that budgeted querying performs \emph{slightly better} than unconstrained one on this task,
although the practical behaviors are hardly distinguishable.
This may be due to budgeted queries' statistical advantages compared to full feedback, which may increase bias of $h_t$ for gradient estimation.
Although small, performance difference appears consistent across all budgets, with a monotonic trend that reinforces its statistical significance.
A deeper study on potential advantages of querying is outside the scope of this paper.

On the other hand,
the impact of optimized behavior does not seem significantly in this case.
With $\beta=0.10$ and policy-aware allocation, the mean normalized policy entropy is $0.99990$ and the mean same-policy ratio $\Phi_t(\rho_t^\star)/\Phi_t(\rho_t^{\mathrm{unif}})$ is $0.99919$, offering little opportunity for improvement over uniform allocation.
This may be partly explained by modest time-variation of the agent set.

\begin{figure}
	\centering
	\includegraphics[width=0.92\textwidth]{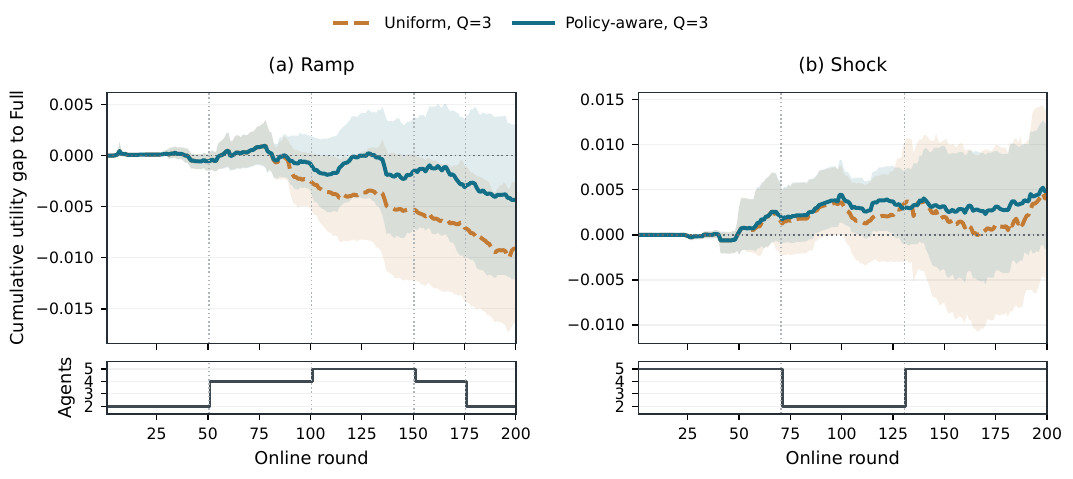}
	\caption{Coverage performance with $Q_t\equiv3$ on log-Gaussian cell-weight fields under ramp (a) and shock (b) agent  schedules.
		Upper panels show mean cumulative-utility gaps to full feedback with $95\%$ confidence intervals regions.
		Lower panels show the number of active agents,
		with vertical dotted lines in both panels marking the changes in agent set.}
	\label{fig:open}
\end{figure}

\subsection{Open-System Response Under Constant Budget}

Next, we hold a fixed budget $Q_t=3$ and test with ramp and shock agent schedules on log-Gaussian cell-weight fields.
In this case,
budgeted methods query about $17\%$ of the gradient coordinates under ramp and $15\%$ under shock schedules.
\autoref{fig:open} shows the cumulative-utility gaps (\emph{not} percentage) with uniform and optimized querying throughout the horizon.
In this case, the latter method yields statistically consistent higher utility, especially with gradually changing agent set.

\section{Conclusion}
\label{sec:conclusion}

This paper presented a KL mirror-ascent method for task allocation in open multi-agent systems.
Deterministic rounding and inverse-probability weighting of counterfactual samples enforce a prescribed budget and preserve unbiased gradient estimation.
We established a regret bound separating the contributions of open system and query budget, respectively,
and proposed a query allocation that minimizes it.
Experiments on cooperative coverage report essentially unchanged performance at a fraction of full-feedback queries,
supporting online learning with cheap compute.
Future work will consider noisy marginal evaluations, neural policies, and deeper compute evaluation.

\subsection*{Generative AI Use}

Generative AI tools were used to refine proofs and write simulation code. All AI-assisted content was reviewed by the authors, who take responsibility for the correctness and integrity of the work.
	
	\newpage
	\bibliography{references}

@inproceedings{liu2026submapg,
	author={Liu, Jing and Yang, Yangyang and Ballotta, Luca and Li, Fangfei and Tang, Yang and Carli, Ruggero},
	title={Submodular Multi-Agent Policy Learning for Online Distributed Task Allocation in Open Multi-Agent Systems},
	booktitle = {Advances in Neural Information Processing Systems},
	year      = {2026},
	note = {[accepted]}
}

@article{liu2026submapl,
	title={Submodular Policy Learning for Distributed Task Allocation in Open Multi-Agent Systems},
	author={Liu, Jing and Ballotta, Luca and Yang, Yangyang and Li, Fangfei and Tang, Yang and Carli, Ruggero},
	journal={arXiv preprint arXiv:2608.14390},
	year={2026}
}

@inproceedings{zhang2025effective,
	author={Zhang, Qixin and Sun, Yan and Jin, Can and Zhang, Xikun and Shu, Yao and Zhao, Puning and Shen, Li and Tao, Dacheng},
	title={Effective Policy Learning for Multi-Agent Online Coordination Beyond Submodular Objectives},
	booktitle = {Advances in Neural Information Processing Systems},
	volume    = {38},
	pages = {164278--164339},
	year      = {2025},
}

@inproceedings{efroni2021confidence,
	author    = {Efroni, Yonathan and Merlis, Nadav and Saha, Aadirupa and Mannor, Shie},
	title     = {Confidence-Budget Matching for Sequential Budgeted Learning},
	booktitle = {Proceedings of the 38th International Conference on Machine Learning},
	series    = {Proceedings of Machine Learning Research},
	volume    = {139},
	pages     = {2937--2947},
	year      = {2021}
}

@inproceedings{li2025constrained,
	author    = {Li, Shaoang and Li, Jian},
	title     = {Constrained Feedback Learning for Non-Stationary Multi-Armed Bandits},
	booktitle = {Advances in Neural Information Processing Systems},
	volume    = {38},
	pages     = {80711--80744},
	year      = {2025},
	doi       = {10.52202/085713-2697}
}

@inproceedings{zhang2019online,
	author    = {Zhang, Mingrui and Chen, Lin and Hassani, Hamed and Karbasi, Amin},
	title     = {Online Continuous Submodular Maximization: From Full-Information to Bandit Feedback},
	booktitle = {Advances in Neural Information Processing Systems},
	volume    = {32},
	pages     = {9206--9217},
	year      = {2019}
}

@inproceedings{zhang2025nearoptimal,
	author    = {Zhang, Qixin and Wan, Zongqi and Yang, Yu and Shen, Li and Tao, Dacheng},
	title     = {Near-Optimal Online Learning for Multi-Agent Submodular Coordination: Tight Approximation and Communication Efficiency},
	booktitle = {International Conference on Learning Representations},
	year      = {2025}
}

@inproceedings{xu2023bandit,
	author    = {Xu, Zirui and Lin, Xiaofeng and Tzoumas, Vasileios},
	title     = {Bandit Submodular Maximization for Multi-Robot Coordination in Unpredictable and Partially Observable Environments},
	booktitle = {Robotics: Science and Systems XIX},
	address   = {Daegu, Republic of Korea},
	month     = jul,
	year      = {2023},
	doi       = {10.15607/RSS.2023.XIX.109}
}

@inproceedings{wan2023bandit,
	author    = {Wan, Zongqi and Zhang, Jialin and Chen, Wei and Sun, Xiaoming and Zhang, Zhijie},
	title     = {Bandit Multi-linear {DR}-Submodular Maximization and Its Applications on Adversarial Submodular Bandits},
	booktitle = {Proceedings of the 40th International Conference on Machine Learning},
	series    = {Proceedings of Machine Learning Research},
	volume    = {202},
	pages     = {35491--35524},
	year      = {2023}
}

@article{gandhi2006dependent,
	author  = {Gandhi, Rajiv and Khuller, Samir and Parthasarathy, Srinivasan and Srinivasan, Aravind},
	title   = {Dependent Rounding and Its Applications to Approximation Algorithms},
	journal = {Journal of the ACM},
	volume  = {53},
	number  = {3},
	pages   = {324--360},
	year    = {2006},
	doi     = {10.1145/1147954.1147956}
}

@article{calinescu2011maximizing,
	author  = {Calinescu, Gruia and Chekuri, Chandra and P{\'a}l, Martin and Vondr{\'a}k, Jan},
	title   = {Maximizing a Monotone Submodular Function Subject to a Matroid Constraint},
	journal = {SIAM Journal on Computing},
	volume  = {40},
	number  = {6},
	pages   = {1740--1766},
	year    = {2011},
	doi     = {10.1137/080733991}
}

@article{xu2023online,
	author  = {Xu, Zirui and Zhou, Hongyu and Tzoumas, Vasileios},
	title   = {Online Submodular Coordination With Bounded Tracking Regret: Theory, Algorithm, and Applications to Multi-Robot Coordination},
	journal = {IEEE Robotics and Automation Letters},
	volume  = {8},
	number  = {4},
	pages   = {2261--2268},
	month   = apr,
	year    = {2023},
	doi     = {10.1109/LRA.2023.3248372}
}

@article{deplano2026optimization,
	author  = {Deplano, Diego and Bastianello, Nicola and Franceschelli, Mauro and Johansson, Karl H.},
	title   = {Optimization and Learning in Open Multi-Agent Systems},
	journal = {IEEE Transactions on Automatic Control},
	volume  = {71},
	number  = {6},
	pages   = {3864--3879},
	year    = {2026},
	doi     = {10.1109/TAC.2026.3653296}
}

@article{cesabianchi2005label,
	author  = {Cesa-Bianchi, Nicol{\`o} and Lugosi, G{\'a}bor and Stoltz, Gilles},
	title   = {Minimizing Regret with Label Efficient Prediction},
	journal = {IEEE Transactions on Information Theory},
	volume  = {51},
	number  = {6},
	pages   = {2152--2162},
	year    = {2005},
	doi     = {10.1109/TIT.2005.847729}
}

@article{nemhauser1978analysis,
	author  = {Nemhauser, George L. and Wolsey, Laurence A. and Fisher, Marshall L.},
	title   = {An Analysis of Approximations for Maximizing Submodular Set Functions---I},
	journal = {Mathematical Programming},
	volume  = {14},
	pages   = {265--294},
	year    = {1978},
	doi     = {10.1007/BF01588971}
}

@inproceedings{Tzoumas16acc-sensorPlacementKalman,
	title = {Sensor placement for optimal {Kalman} filtering: {Fundamental} limits, submodularity, and algorithms},
	shorttitle = {{sensorPlacementKalman}},
	doi = {10.1109/ACC.2016.7524914},
	urldate = {2025-02-26},
	booktitle = {2016 {American} {Control} {Conference} ({ACC})},
	author = {Tzoumas, V. and Jadbabaie, A. and Pappas, G. J.},
	month = jul,
	year = {2016},
	note = {acc},
	pages = {191--196},
}

@article{beck2003mirror,
	title={Mirror descent and nonlinear projected subgradient methods for convex optimization},
	author={Beck, Amir and Teboulle, Marc},
	journal={Operations Research Letters},
	volume={31},
	number={3},
	pages={167--175},
	year={2003},
	publisher={Elsevier}
}
	
	\clearpage
	\setlength{\abovedisplayskip}{5pt plus 2pt minus 1pt}
	\setlength{\belowdisplayskip}{5pt plus 2pt minus 1pt}
	\setlength{\abovedisplayshortskip}{2pt plus 1pt minus 1pt}
	\setlength{\belowdisplayshortskip}{3pt plus 1pt minus 1pt}
	\appendix

\section*{Appendix: Table of Contents}
\begingroup
\makeatletter\@starttoc{apx}\makeatother
\endgroup
\medskip

\section{Assumption~\ref{ass:stage}}
\label{app:ass-utility}
\addcontentsline{apx}{section}{\protect\numberline{\thesection}Assumption~\ref{ass:stage}}

Set function $F_t(\cdot;s_t):2^{\Omega_t}\to\mathbb{R}$ is:

\begin{itemize}
	\item Normalized: $F_t(\varnothing;s_t)=0$;
	\item Monotone: $F_t(B;s_t)\le F_t(C;s_t)$ for every $B\subseteq C\subseteq\Omega$;
	\item submodular: $F_t(e\mid B;s_t)\ge F_t(e\mid C;s_t)$ for $B\subseteq C$ and $e\notin C$.
\end{itemize}

\section{Proof of Theorem~\ref{thm:query-regret}}
\label{app:proof-main-regret}
\addcontentsline{apx}{section}{\protect\numberline{\thesection}Proof of Theorem~\ref{thm:query-regret}}

Before proving the main theorem,
we introduce a few auxiliary lemmas.

\begin{lemma}[Restricted PME inequality]\label{prop:restricted}
	Under Assumption~\ref{ass:stage}, for any $x,y\in\cF_t$,
	\begin{equation}\label{eq:restricted-pme}
		\frac12\widetilde f_t(y;s_t)-\widetilde f_t(x;s_t)
		\le
		\frac12\langle\nabla\widetilde f_t(x;s_t),y-x\rangle.
	\end{equation}
\end{lemma}

\begin{proof}
	Draw $A=\{e_i:i\in\cN_t\}\sim\mathcal D_t(x)$ and
	$Y=\{u_i:i\in\cN_t\}\sim\mathcal D_t(y)$ independently.  Monotonicity
	and a telescoping expansion of $F_t(A\cup Y;s_t)-F_t(A;s_t)$ give
	\[
	F_t(Y;s_t)-F_t(A;s_t)
	\le
	\sum_{i\in\cN_t}F_t(u_i\mid A^{-i};s_t).
	\]
	Add the elements of $Y$ in a fixed agent order.  If $u_i\ne e_i$,
	the set present before adding $u_i$ contains $A^{-i}$.  Submodularity
	therefore bounds the increment by $F_t(u_i\mid A^{-i};s_t)$.  If $u_i=e_i$, the union already contains
	that element, so the increment is zero and the same upper bound follows
	from monotonicity.
	
	Normalization and a telescoping expansion of $F_t(A;s_t)$ in the same
	agent order give
	\[
	\sum_{i\in\cN_t}F_t(e_i\mid A^{-i};s_t)\le F_t(A;s_t).
	\]
	Subtracting this bound from the preceding one yields, for every realization,
	\[
	F_t(Y;s_t)-2F_t(A;s_t)
	\le
	\sum_{i\in\cN_t}
	\bigl[
	F_t(u_i\mid A^{-i};s_t)-F_t(e_i\mid A^{-i};s_t)
	\bigr].
	\]
	Take expectation over the independent draws $A$ and $Y$.  For every
	coordinate, differentiating the extension on $\mathcal P_t$ gives
	\[
	\frac{\partial\widetilde f_t(x;s_t)}{\partial x_{i,a}}
	=
	\E_{A^{-i}\sim\mathcal D_t^{-i}(x)}
	[F_t((i,a)\mid A^{-i};s_t)].
	\]
	Consequently,
	\[
	\widetilde f_t(y;s_t)-2\widetilde f_t(x;s_t)
	\le
	\langle\nabla\widetilde f_t(x;s_t),y-x\rangle.
	\]
	Division by two proves the result.
\end{proof}

Note that setting $x=x_t$ and $y=x_t^\textnormal{OPT}$ provides the bound underlying the factor $\nicefrac{1}{2}$ in the definition of approximate dynamic regret in~\eqref{eq:dynamic-regret}.

For subsequent analysis, let $\mathcal H_t=\sigma(\mathcal I_t^-,F_t)$ and $\mathcal G_t=\sigma(\mathcal H_t,A_t)$ respectively denote the $\sigma$-algebras generated by available information and utility at round $t$ and, additionally, the joint action samples at round $t$.

\begin{lemma}[Gradient estimator moments]
	\label{lem:sparse-feedback}
	Assume that $\rho_{i,a,t}>0$ and that $A_t$, conditional on $\mathcal H_t$, follows $\mathcal D_t(x_t)$. Then the estimator in \eqref{eq:sparse-estimator} satisfies
	\begin{align}
		\E[\widehat g_{i,a,t}\mid\mathcal G_t]&=h_{i,a,t},\label{eq:conditional-unbiased}\\
		\E[\widehat g_{i,a,t}^{\,2}\mid\mathcal G_t]&=\frac{h_{i,a,t}^{\,2}}{\rho_{i,a,t}}.\label{eq:conditional-moment}
	\end{align}
	Consequently,
	\begin{equation}
		\E[\widehat g_t\mid\mathcal H_t]=\nabla\widetilde f_t(x_t;s_t).
		\label{eq:pme-unbiased}
	\end{equation}
\end{lemma}

\begin{proof}
	Conditional on $\mathcal G_t$, the marginal gain $h_{i,a,t}$ and inclusion probability $\rho_{i,a,t}$ are fixed, and $\E[Z_{i,a,t}\mid\mathcal G_t]=\Pr(Z_{i,a,t}=1\mid\mathcal G_t)=\rho_{i,a,t}$. Substitution into \eqref{eq:sparse-estimator} gives \eqref{eq:conditional-unbiased} and identity $Z_{i,a,t}^2=Z_{i,a,t}$ gives \eqref{eq:conditional-moment}. Taking the conditional expectation of \eqref{eq:conditional-unbiased} with respect to $\mathcal H_t$ and applying \eqref{eq:marginal-gradient-link} yields \eqref{eq:pme-unbiased}.
\end{proof}

The next lemma uses the entropic mirror-descent framework~\citep{beck2003mirror}, with a second-order term suited to the nonnegative marginal-gain estimates.

\begin{lemma}[Second-order KL mirror-ascent bound]
	\label{lem:entropic}
	Let $x,u$ belong to the same product of categorical simplexes, with $x>0$. Given $g\ge0$ and $\eta>0$ such that $\eta g_{i,a}\le1$ for all $(i,a)$, let $x^+$ be the KL mirror-ascent update of $x$ with direction $g$ as in \eqref{eq:mirror-update}. Then
	\begin{equation}
		\langle g,u-x\rangle\le{}\frac{\KL(u\Vert x)-\KL(u\Vert x^+)}{\eta} + (e-2)\eta\sum_{i,a}x_{i,a}g_{i,a}^2.
		\label{eq:second-order-step}
	\end{equation}
\end{lemma}

\begin{proof}
	For an agent $i$, let $M_i=\sum_a x_{i,a}\exp(\eta g_{i,a})$. The mirror update gives
	\begin{equation*}
		\KL(u_i\Vert x_i)-\KL(u_i\Vert x_i^+)=\eta\langle u_i,g_i\rangle-\log M_i.
	\end{equation*}
	Since $e^z\le1+z+(e-2)z^2$ for $z\in[0,1]$ and $\log(1+r)\le r$ for $r>-1$,
	\begin{equation*}
		\log M_i\le\eta\langle x_i,g_i\rangle+(e-2)\eta^2\sum_a x_{i,a}g_{i,a}^2.
	\end{equation*}
	Replacing this inequality into the KL identity, summing over the agents, and dividing by $\eta$ proves \eqref{eq:second-order-step}.
\end{proof}

By Assumption~\ref{ass:stage} and the inclusion-probability lower bound~\eqref{eq:rho-range}, it follows $0\le\widehat g_{i,a,t}\le \nicefrac{L}{\rho_{\min}}$.
Hence, the condition of Lemma~\ref{lem:entropic} holds for every realization when
\begin{equation}
	0<\eta\le\frac{\rho_{\min}}{L}.
	\label{eq:stepsize-condition}
\end{equation}

\paragraph*{Proof of Theorem~\ref{thm:query-regret}}

The restricted first-order inequality \eqref{eq:restricted-pme} gives
\begin{equation*}
	\frac12\widetilde f_t(x_t^\star;s_t)-\widetilde f_t(x_t;s_t)\le\frac12\langle\nabla\widetilde f_t(x_t;s_t),x_t^\star-x_t\rangle.
\end{equation*}
The comparator $x_t^\star$ and the current policy $x_t$ are $\mathcal H_t$-measurable. By \eqref{eq:pme-unbiased}, the expectation of the right-hand side equals $\frac12\E[\langle\widehat g_t,x_t^\star-x_t\rangle]$. Applying Lemma~\ref{lem:entropic} to $x_t\mapsto\bar x_{t+1}$, followed by \eqref{eq:conditional-moment}, yields
\begin{equation}\label{eq:proof-thm}
	\begin{aligned}
		\E\!\left[\frac12\widetilde f_t(x_t^\star;s_t)-\widetilde f_t(x_t;s_t)\right]
	&\le\frac{\E[\KL(x_t^\star\Vert x_t)]-\E[\KL(x_t^\star\Vert\bar x_{t+1})]}{2\eta}\\
	&+\frac{(e-2)\eta}{2}\E\!\left[\sum_{(i,a)\in\Omega_t}\frac{x_{i,a,t}h_{i,a,t}^{\,2}}{\rho_{i,a,t}}\right].
	\end{aligned}
\end{equation}
Set $\Phi_t=\KL(x_t^\star\Vert x_t)$ and $\Gamma_t=\KL(x_t^\star\Vert\bar x_{t+1})$. The KL terms satisfy the telescoping identity
\begin{equation*}
	\sum_{t=1}^T(\Phi_t-\Gamma_t)=\Phi_1-\Gamma_T+\sum_{t=1}^{T-1}(\Phi_{t+1}-\Gamma_t).
\end{equation*}
Since $\Gamma_T\ge0$, \eqref{eq:open-variation} implies $\E[\sum_{t=1}^T(\Phi_t-\Gamma_t)]\le\E[\Phi_1]+\mathcal V_T^{\mathrm{open}}$. Summing the one-step inequalities and combining the last expectation in~\eqref{eq:proof-thm} with Assumption~\ref{ass:stage} proves the claim.

\section{Proof of Proposition~\ref{prop:allocation}}
\label{app:proof-allocation}
\addcontentsline{apx}{section}{\protect\numberline{\thesection}Proof of Proposition~\ref{prop:allocation}}

Since all active policy probabilities are positive, the objective in \eqref{eq:allocation-program} is strictly convex on its feasible set, which is nonempty and compact under the budget condition.
For an interior coordinate, the KKT stationarity condition yields $-x_{i,a,t}/\rho_{i,a,t}^2+\lambda=0$, hence $\rho_{i,a,t}=\sqrt{x_{i,a,t}/\lambda}$. KKT conditions and complementary slackness yield $\rho_{i,a,t}^{\star}=\operatorname{clip}_{[\rho_{\min,t},1]}\!\left(\tau_t\sqrt{x_{i,a,t}}\right)$.
The scalar $\tau_t=\lambda^{-1/2}$ is selected to satisfy the budget equality, and strict convexity ensures a unique minimizer.
At the extreme feasible budgets, the allocation is uniquely determined by the constraints and satisfies the clipping condition. Equalities~\eqref{eq:unclipped-rho} follow directly in the absence of clipping.

\section{Proof of Corollary~\ref{cor:regret}}
\label{app:proof-cor-regret}
\addcontentsline{apx}{section}{\protect\numberline{\thesection}Proof of Corollary~\ref{cor:regret}}

Uniform inclusion probabilities $\rho_{i,a,t}^{\mathrm{unif}}=\nicefrac{Q_t}{d_t}$ are feasible for problem~\eqref{eq:allocation-program}.
Since categorical agent policies satisfy $\sum_{(i,a)\in\Omega_t}x_{i,a,t}=|\mathcal N_t|$,
optimality $\Phi_t(\rho_t^\star)\le\Phi_t(\rho_t^{\mathrm{unif}})$ implies
\begin{equation}
	\sum_{(i,a)\in\Omega_t}\frac{x_{i,a,t}}{\rho_{i,a,t}^\star}\le\frac{d_t|\mathcal N_t|}{Q_t}
	\label{eq:global-budget-rate}
\end{equation}
and the claim follows by replacing bound~\eqref{eq:global-budget-rate} into~\eqref{eq:main-regret}.
	
\end{document}